\documentclass[runningheads]{llncs}

\usepackage[final,year=2026]{eccv}
\usepackage{eccvabbrv}
\usepackage{graphicx}
\usepackage{booktabs}
\usepackage{placeins}
\usepackage{amsmath}
\usepackage{hyperref}
\hypersetup{
    hidelinks,
    pdftitle={How Merge-Tolerant Are Vision Transformers for Wheat Phenotyping?},
    pdfauthor={Simon Rav\'e, Pejman Rasti, and David Rousseau}
}
\usepackage{xcolor}
\usepackage{tikz}
\usetikzlibrary{positioning,fit,arrows.meta,shapes.geometric,calc,backgrounds}

\begin{document}

\title{How Merge-Tolerant Are Vision Transformers for Wheat Phenotyping?}
\titlerunning{Merge-Tolerant Vision Transformers}

\author{Simon Rav\'e\inst{1}\thanks{Corresponding author.} \and Pejman Rasti\inst{1,2} \and David Rousseau\inst{1,2}\\
\email{\{simon.rave,pejman.rasti,david.rousseau\}@univ-angers.fr}}
\authorrunning{S. Rav\'e et al.}
\institute{LARIS, University of Angers, Angers, France
\and UMR INRAE-IRHS, Angers, France}

\maketitle

\begin{abstract}
Vision-based wheat phenotyping requires repeated measurements under deployment constraints, from growth-stage recognition to wheat-head counting and organ segmentation. Plain Vision Transformers (ViTs) provide a common architecture for these tasks, but quadratic attention limits high-throughput and edge inference. Training-free token merging is attractive because it can be inserted into trained models without retraining. We provide a systematic benchmark of ToMe and Mutual Pair Merging across growth-stage classification, wheat-head detection, and wheat-organ segmentation, measuring task quality, throughput, token count, and peak GPU memory, with additional Raspberry Pi 5 measurements. The benchmark reveals a clear hierarchy: classification is highly merge-tolerant, while detection and segmentation are constrained by repeated instances, thin organs, dense boundaries, reconstruction, and runtime overhead. Optimized attention backends can erase apparent speedups, so deployment value must be profiled on the target runtime rather than inferred from token count.

\keywords{Plant phenotyping \and Vision transformers \and Token merging \and Efficient inference}
\end{abstract}

\section{Introduction} 

\begin{figure}
    \centering
    \resizebox{\linewidth}{!}{%
        \input{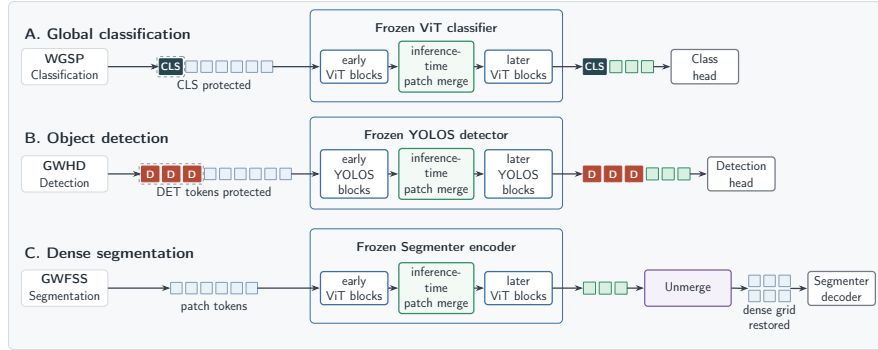}%
    }
    \caption{Overview of our merging protocol. We benchmark classification, segmentation, and detection using plain Vision Transformers~\cite{dosovitskiy2021vit}. After training and freezing each model, we apply ToMe~\cite{bolya2023tome} or MPM~\cite{rave2026mpm} only at inference. We evaluate multiple backbone sizes and report task quality, token count, latency, and throughput on an NVIDIA RTX PRO 6000 system and a Raspberry Pi 5, with peak memory reported for GPU runs.}
    \label{fig:main}
\end{figure}

Image-based plant phenotyping has moved from isolated image analysis toward repeated measurement in fields, greenhouses, and breeding nurseries. Cameras on fixed stations, UAVs, robots, and low-cost edge devices produce visual streams across sites, growth stages, and acquisition platforms under varying device budgets. A breeder may need the growth stage of a wheat canopy, the number and location of wheat heads, or a pixel-level separation of leaves, stems, and spikes. A model that is accurate but too slow, too memory-hungry, or too dependent on workstation hardware may be difficult to use in field robots or low-cost phenotyping systems~\cite{joice2025raspberry}. Recent work has therefore explored general visual representations and compact models for plant recognition~\cite{murphy2024deep,chen2023adapting,elamine2023embedded}.

Wheat images contain strong visual redundancy (\cref{fig:similarity}). Repeated biological structures create both an opportunity and a risk: canopy evidence may be compressed, whereas small instances and boundaries may be merged away. Plain ViTs provide the same sequence interface for classification, segmentation, and detection~\cite{dosovitskiy2021vit,strudel2021segmenter,fang2021yolos}, but self-attention scales poorly with sequence length while crop images often require high resolution. The question is whether their token sequences can be compressed without losing phenotype-relevant evidence.

Training-free token merging addresses this deployment pressure by aggregating similar tokens inside an already trained transformer. ToMe showed that this intervention can improve throughput without retraining~\cite{bolya2023tome}. MPM emphasizes end-to-end latency and dense-feature reconstruction~\cite{rave2026mpm}. Yet generic computer-vision conclusions do not directly transfer to plant phenotyping: a merger that is harmless for classification may be harmful when the signal is one wheat head among many, a narrow stem, or a leaf boundary. The answer is also hardware-dependent because merge construction and reconstruction can dominate on optimized GPUs even when shorter sequences help embedded processors~\cite{dao2024flashattention2}.

We therefore ask when an already trained ViT can be accelerated by merging tokens and when merging destroys the biological signal of interest. We study WGSP growth-stage classification~\cite{shen2024gspai}, GWHD 2021 wheat-head detection~\cite{david2021gwhd}, and GWFSS wheat-organ segmentation~\cite{wang2025gwfss}, using a ViT classifier, Segmenter, and YOLOS across multiple backbone sizes. We report primary and secondary task metrics together with measured GPU and Raspberry Pi 5 behavior. Classification tolerates substantial token reduction, detection provides limited practical gains in the tiled pipeline, and segmentation degrades under aggressive merging and dense reconstruction. Our contributions are:
\begin{enumerate}
    \item We provide a benchmark of training-free token merging for ViT-based wheat phenotyping across classification, detection, and segmentation under a unified inference-time protocol.
    \item We compare ToMe and MPM on plain ViT-based backbones and report task quality together with measured throughput, latency, final token count, and peak GPU memory, so that accuracy tolerance is separated from deployment usefulness.
    \item We analyze why merge tolerance differs across plant tasks, including merge locality, dense reconstruction cost, adaptive padding, attention backends, and Raspberry Pi 5 edge behavior, and derive task- and runtime-specific deployment guidance.
    \item The code is available at \url{https://forge.inrae.fr/imhorphen/wheat-token-merging}.
\end{enumerate}

\section{Related Work}

\paragraph{Vision Transformers for Plant Vision.}
Vision Transformers represent an image as patch tokens processed by self-attention~\cite{dosovitskiy2021vit}. The abstraction extends beyond classification: Segmenter applies a plain ViT encoder with a mask-transformer decoder~\cite{strudel2021segmenter}, while YOLOS formulates detection with a minimal ViT-style sequence interface~\cite{fang2021yolos}. This common interface matters because token merging acts on the sequence rather than a task-specific convolutional pyramid. Plant-vision work has adapted foundation models and compact networks to recognition and measurement~\cite{murphy2024deep,chen2023adapting,elamine2023embedded}. We instead ask how much of an already trained plant ViT sequence can be merged without losing task utility.

\paragraph{Token Merging.}
Token reduction can prune tokens~\cite{rao2021dynamicvit,liang2022evit,yin2022avit,ryoo2021tokenlearner,fayyaz2022ats} or merge them. We focus on merging because it preserves information by aggregation rather than deletion. Token Pooling formulates aggregation as clustering~\cite{marin2023tokenpooling}. ToMe pairs similar tokens inside a pretrained ViT without retraining~\cite{bolya2023tome}, whereas MPM uses mutual nearest-neighbor pairs and records mappings for dense reconstruction~\cite{rave2026mpm}. Prior comparisons show that accuracy--speed behavior cannot be inferred from token count alone~\cite{haurum2023which}. The trade-off is also task-dependent~\cite{lu2023cts,norouzi2024algm,bergner2025tokencropr}: classification often tolerates coarse reduction, while dense prediction requires recoverable spatial evidence. We therefore evaluate both regimes using task quality and measured runtime rather than token count as a proxy.

\paragraph{Wheat Phenotyping Benchmarks and Deployment.}
Wheat phenotyping tasks differ in the spatial evidence they require. WGSP/GSP-AI uses canopy imagery for growth-stage recognition~\cite{shen2024gspai}, GWHD evaluates head detection and counting across diverse conditions~\cite{david2021gwhd,david2023gwhdchallenges}, and GWFSS requires semantic separation of leaves, stems, and spikes~\cite{wang2025gwfss}. These datasets expose global canopy structure, many small repeated objects, and dense boundary-sensitive organs. Embedded benchmarks and Raspberry Pi studies further motivate operation on constrained agricultural hardware~\cite{elamine2023embedded,joice2025raspberry}. To our knowledge, prior plant-vision work has not evaluated training-free token merging across all three regimes, multiple plain-ViT sizes, and both edge and workstation hardware.

\section{Method}
\label{sec:method}

\subsection{Benchmark Design}

We study token merging as a deployment intervention, not as a new plant architecture. For each task, we first train a plain ViT-based model without token merging. We then freeze all weights and insert a training-free token merger only at inference time. Thus, for a fixed checkpoint, any change in task metrics or deployment measurements comes from the modified token sequence and from the overhead of the merge operator itself.

Let $x \in \mathbb{R}^{H \times W \times 3}$ be an RGB image. A ViT with patch size $P$ maps $x$ to a sequence of image tokens $Z_0 \in \mathbb{R}^{N_0 \times d}$, where $N_0 = HW/P^2$ for a square image divisible by $P$~\cite{dosovitskiy2021vit}. Depending on the task, the sequence may also contain protected tokens $U_l$, such as a classification token or detection tokens. At a selected transformer layer $\ell$, a merger replaces only the image-token sequence,
\begin{equation}
    X_\ell = [U_\ell, Z_\ell]
    \quad \longrightarrow \quad
    \widetilde{X}_\ell = [U_\ell, M_\ell(Z_\ell)],
\end{equation}
where $M_\ell$ returns fewer image tokens. Protected tokens are never merged. The task head is left unchanged.

We define merge tolerance by two quantities. The first is the task-performance drop relative to the unmerged checkpoint. The second is measured speedup,
\begin{equation}
    s = \frac{T_{\mathrm{base}}}{T_{\mathrm{merge}}},
\end{equation}
where $T$ is wall-clock inference time under the same batch size, precision, device, and attention backend. A merging setting is deployment-useful only if it mostly preserves the task metric while giving $s>1$ in the measured path.

\subsection{Tasks and Datasets}

The benchmark covers one global, one instance-level, and one dense prediction task. This triplet mirrors a common phenotyping progression from coarse crop-state assessment to object-level counting and dense organ measurement. It also makes merge tolerance interpretable because the tasks differ not only by metric, but also by the spatial scale and identity of the biological evidence they require. The merging scheme for each task is summarized in \cref{fig:main}.

\begin{table}[ht]
\centering
\small
\caption{Tasks, dataset splits, and metrics. WGSP and GWHD list training / validation / test, GWFSS lists training / evaluation.}
\label{tab:task_protocol}
\resizebox{\textwidth}{!}{\begin{tabular}{llll}
\hline
Task & Dataset & Split & Metrics \\
\hline
Classification & WGSP/GSP-AI~\cite{shen2024gspai} &
36,440 / 5,206 / 10,412 &
Top-1, macro-F1 \\
Detection & GWHD 2021~\cite{david2021gwhd} &
3,605 / 1,448 / 1,334 &
mAP, AP$_{50}$, count MAE, count $R^2$ \\
Segmentation & GWFSS~\cite{wang2025gwfss} &
876 / 220 &
mIoU, per-class IoU, boundary F1 \\
\hline
\end{tabular}}
\end{table}

\begin{figure}[t!]
    \centering
    \includegraphics[width=\linewidth]{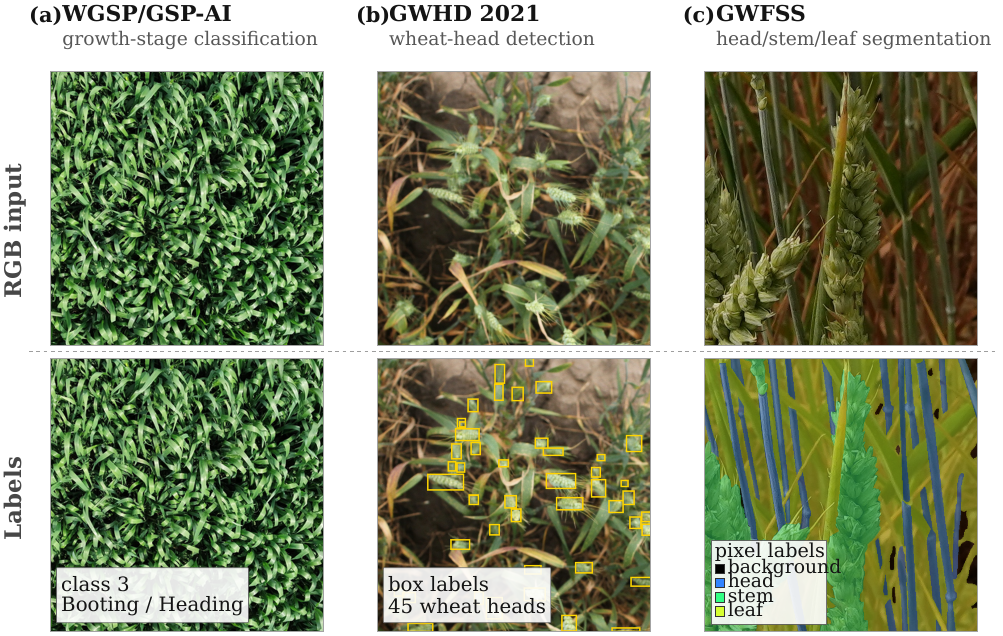}
    \caption{The three wheat datasets: WGSP/GSP-AI growth-stage classification (left), GWHD 2021 head detection (center), and GWFSS organ segmentation (right). They require global canopy evidence, localization of repeated instances, and dense separation of boundary-sensitive organs, respectively.}
    \label{fig:datasets-overview}
\end{figure}

Representative samples appear in \cref{fig:datasets-overview}. WGSP assigns one Zadoks growth-stage label to a canopy image, so evidence may be pooled over broad plant regions. GWHD instead requires the model to retain the location and identity of individual wheat heads despite their repeated appearance and variable density. GWFSS requires pixel-level separation of Background, Head/Spike, Stem, and Leaf, including narrow organs and interfaces between adjacent classes. Its provided 220-image evaluation split is also used for checkpoint selection and final reporting, so it is not an independently held-out test estimate. The primary metrics are Top-1, mAP, and mIoU. Macro-F1, AP$_{50}$, count MAE, count $R^2$, per-class IoU, and boundary F1 test whether an apparently tolerant operating point conceals class-, counting-, or boundary-specific damage.

\subsection{Backbones and Training}

For WGSP, pretrained \texttt{timm} ViT-T/16, ViT-S/16, and ViT-B/16 classifiers~\cite{dosovitskiy2021vit} are fine-tuned for 50 epochs with AdamW~\cite{loshchilov2019decoupled}, batch size 256, weight decay $0.05$, and seed 1337. The constant learning rates are $4\!\times\!10^{-4}$ for Tiny and Small and $2\!\times\!10^{-4}$ for Base. Cross-entropy training uses random resized crops ($[0.7,1.0]$), horizontal flips, and RandAugment (two operations, magnitude 9)~\cite{cubuk2020randaugment}. The best validation macro-F1 checkpoint is evaluated at $224^2$.

For GWFSS~\cite{wang2025gwfss}, Segmenter combines the corresponding pretrained \texttt{timm} ViT-T/16, ViT-S/16, or ViT-B/16 encoder with a two-layer mask-transformer decoder~\cite{strudel2021segmenter}. All variants use AdamW for 80 epochs, weight decay $0.05$, cosine decay after a five-epoch warm-up, weighted cross-entropy $(0.57,1.78,1.17,0.49)$, bfloat16, and seed 1337. Tiny uses a learning rate of $2\!\times\!10^{-4}$ and a batch size of 128. Small and Base use $10^{-4}$ with batch sizes of 64 and 32. Augmentation comprises random resized crops to $512^2$ ($[0.45,1.0]$), horizontal and vertical flips, rotations up to $15^\circ$, color jitter, grayscale conversion, and Gaussian blur. The best evaluation-split mIoU checkpoint is reported at $512^2$.

For GWHD, we use pretrained YOLOS models~\cite{fang2021yolos}. Each model is fine-tuned first on full images for 30 epochs at $640^2$ (AdamW, learning rate $5\!\times\!10^{-5}$, weight decay $10^{-4}$, batch 128, two warm-up epochs, seed 1337), then on tiles with 25\% overlap and a size of $512$ pixels for six epochs at $640^2$ ($10^{-5}$, batch 64, one warm-up epoch) and four at $768^2$ ($5\!\times\!10^{-6}$, batch 40, half-epoch warm-up), with cosine decay, bfloat16, gradient clipping at 1.0, and seed 20260615. Random resized crops are used only in the initial full-image stage. Flips, color jitter, grayscale conversion, and blur are used throughout. Checkpoints are selected by validation mAP. Validation AP$_{50}$, then mAP, selects the tiled protocol, while count MAE, then RMSE, selects the count threshold. The checkpoint fine-tuned at $768^2$ is evaluated at $640^2$ on nine tiles per image, with NMS IoU $0.3$ and count threshold $0.9$.

The resolutions reflect the pipelines' spatial demands: $224^2$ is native to the classifiers, $512^2$ retains a dense organ grid, and the selected detector evaluates tiles at $640^2$. With patch size 16 these yield 196, 1024, and 1600 image tokens. YOLOS adds 100 detection tokens and nine tiled forward passes. Resolution and tiling therefore confound cross-task throughput comparisons.

\subsection{Training-Free Token Merging}

We compare ToMe~\cite{bolya2023tome} and MPM~\cite{rave2026mpm} on fixed checkpoints without learned parameters. They represent two practical controls: ToMe exposes a fixed per-layer reduction, whereas MPM forms an input-dependent number of mutual pairs at selected layers. This contrast tests a fixed token-removal budget against adaptive aggregation under the same inference-only protocol.

\paragraph{ToMe.}
ToMe removes a fixed number $r$ of image tokens at each selected transformer block. Similarity is computed from normalized attention-key features before tokens are paired and merged. Unless specified otherwise, the schedule is applied in every block. We sweep $r$ with and without proportional attention (\cref{tab:locked-merging-settings}). Stored unmerge maps can be applied in reverse order when a downstream decoder requires the original grid.

\paragraph{MPM.}
At each insertion layer, MPM computes cosine similarity, forms mutual nearest-neighbor pairs, and replaces each accepted pair by its average. Unpaired tokens are retained. MPM has no continuous keep ratio: the discrete schedule $L$ controls both compression and merge-map overhead. Because the retained count is input-dependent, we report effective final tokens and the padding required for batched execution.

\subsection{Reconstruction and Protected Tokens}

Segmenter's decoder expects a dense image-token grid. We reconstruct it after the final ViT block and before the unchanged mask-transformer decoder, using MPM's recorded original-to-merged mapping or ToMe's reverse unmerge maps. Reconstruction preserves tensor compatibility but duplicates merged features at their source positions. It does not recover removed spatial detail.

For YOLOS, only image patch tokens are merged. The 100 detection tokens remain protected and are concatenated with the shortened patch sequence at each modified block. The unchanged head reads them after the final block, testing whether patch-token compression accelerates their interaction without removing evidence required for localization and counting.

\subsection{Evaluation and Profiling}

Each checkpoint is evaluated unmerged and with ToMe or MPM using identical trained weights. Task and timing metrics for a comparison therefore refer to the same frozen model.

Latency is measured in native PyTorch evaluation mode without gradients and includes merge construction, aggregation, reconstruction when required, and the unchanged task head. The scope is the classification forward benchmark, segmentation metric pass, or full tiled detection evaluation. It is fixed within each task but differs across tasks. Detection count $R^2=1-\sum_i(\hat c_i-c_i)^2/\sum_i(c_i-\bar c)^2$ is computed over per-image head counts at the count-MAE score threshold. Peak allocated CUDA memory is reported in decimal GB after resetting PyTorch's counter. It describes framework allocations in the measured path, not total workstation memory, and is comparable only within a task.

We report workstation measurements on an NVIDIA RTX PRO 6000 Blackwell GPU and batch-one edge measurements on a Raspberry Pi 5. The edge experiment measures latency and throughput. We separately study batch size because adaptive merging may require padding.

Finally, we analyze task-specific Pareto curves. A configuration is useful only when it improves measured wall-clock performance while preserving the task metric to an application-appropriate tolerance. This distinction is central for dense prediction, where merge construction, padding, tiling, and grid reconstruction can dominate the nominal saving from shorter sequences.

\section{Results}

\begin{table}[t]
\centering
\footnotesize
\setlength{\tabcolsep}{8pt}
\renewcommand{\arraystretch}{0.95}
\caption{Selected token-merging settings. ToMe uses \(r\) and optional proportional attention. MPM uses insertion layers \(L\).}
\label{tab:locked-merging-settings}
\begin{tabular}{ccl}
\toprule
\multicolumn{2}{c}{ToMe} & \multicolumn{1}{c}{MPM} \\
\cmidrule(lr){1-2}\cmidrule(lr){3-3}
\(r\) & Prop. attn. & \multicolumn{1}{c}{Layers \(L\)} \\
\midrule
8 & Yes & \(\{2\}\) \\
8 & No & \(\{6\}\) \\
16 & Yes & \(\{8\}\) \\
16 & No & \(\{10\}\) \\
24 & Yes & \(\{0,3\}\) \\
24 & No & \(\{1,3\}\) \\
32 & Yes & \(\{2,5\}\) \\
32 & No & \(\{2,7\}\) \\
48 & Yes & \(\{6,9\}\) \\
48 & No & \(\{3,6,9\}\) \\
64 & Yes & \(\{0,2,4,6,8\}\) \\
64 & No & \(\{0,2,4,6,8,10\}\) \\
\bottomrule
\end{tabular}
\end{table}

\begin{table*}[t]
\centering
\small
\setlength{\tabcolsep}{4.2pt}
\caption{Matched final-token comparison on Base backbones at batch size 32 with common within-task hardware, backend, and forward-timing scope. Tokens denotes the mean final total. $\Delta$ is relative to Full. Detection FPS aggregates nine tiled forward passes.}
\label{tab:matched-token-comparison}
\begin{tabular}{llrrrr}
\toprule
Task & Method & Tokens & Metric (\%) & $\Delta$ pp & FPS \\
\midrule
Classification & ToMe $r=8$ & 101.0 & 99.19 & +0.05 & 2975.7 \\
Classification & MPM $L=\{3,6,9\}$ & 100.6 & 99.17 & +0.03 & 3189.9 \\
\addlinespace
Segmentation & ToMe $r=48$ & 449.0 & 62.46 & -1.17 & 266.2 \\
Segmentation & MPM $L=\{3,6,9\}$ & 457.7 & 62.52 & -1.11 & 284.0 \\
\addlinespace
Detection & ToMe $r=32$ & 1317.0 & 26.60 & +1.10 & 5.471 \\
Detection & MPM $L=\{2\}$ & 1305.2 & 22.18 & -3.32 & 5.596 \\
\bottomrule
\end{tabular}
\end{table*}

\cref{tab:matched-token-comparison} provides a post-hoc comparison of the closest available ToMe--MPM final-token pairs in the evaluated sweep. At approximately 101 tokens, the two classification methods differ by only 0.02 Top-1 points and MPM is faster. At approximately 450 tokens, the two segmentation methods differ by 0.06 mIoU points, again with higher MPM throughput. Detection behaves differently: ToMe and MPM retain approximately 1,310 tokens but differ by 4.42 mAP points. Equal token count therefore does not imply equal information retention, particularly when the merger and insertion schedule modify representation dynamics differently. Positive deltas remain descriptive fixed-checkpoint estimates.

\begin{figure}
    \centering
    \includegraphics[width=\linewidth]{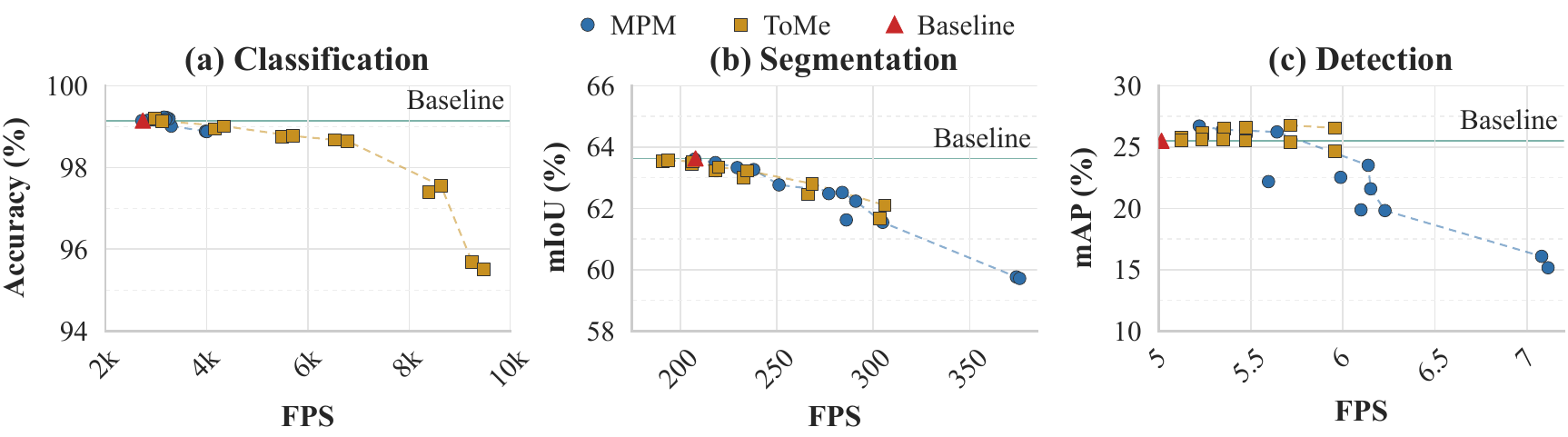}
    \caption{GPU performance--throughput trade-offs on Base backbones. Each point is a merging configuration. The x-axis is FPS and the y-axis is Top-1, mAP, or mIoU. Dashed lines show method-specific Pareto frontiers.}
    \label{fig:locked_three_task_fps_metric_summary}
\end{figure}

We report the main throughput--accuracy trade-offs in \cref{fig:locked_three_task_fps_metric_summary} for the Base backbones on GPU with batch size \(B=32\). The evaluated ToMe and MPM configurations are listed in \cref{tab:locked-merging-settings}. Across the three tasks, the Pareto curves show that merge tolerance follows the spatial scale of the phenotyping signal. Growth-stage classification is the clearest positive case: the model can discard many patch tokens while preserving canopy-level information. Segmentation is intermediate: moderate merging can preserve organ masks, but the mIoU loss grows as token reduction becomes aggressive. Detection is the most constrained in practice: mild merging can preserve or numerically increase mAP, but stronger compression harms localization and the tiled inference pipeline limits wall-clock gains. For GWHD, the original images are high-resolution and wheat heads are small, so each image is divided into nine tiles before inference. This makes detection substantially slower than the classification and segmentation settings and makes throughput gains harder to realize end to end.

Representative operating points appear in \cref{tab:selected-base-pareto-results}. They preserve the primary metric while improving or approximately preserving throughput, rather than claiming exhaustive optima. For classification, ToMe and MPM reach \(1.59\times\) and \(1.47\times\) speedups for Top-1 losses of 0.13 and 0.26 points. They end with 11 and 79 tokens, reinforcing that token count is not itself semantic information. For segmentation, \(1.33\times\) and \(1.16\times\) speedups cost 0.82 and 0.36 mIoU points. Detection mAP increases numerically by 1.27 and 1.21 points, but speedups are only \(1.11\times\) and \(1.02\times\). Additionally, we report secondary metrics for the main configurations in \cref{tab:selected-secondary-memory}. We see that peak memory is higher when merging for segmentation. This is due to the merge maps being stored for reconstruction. For detection, AP$_{50}$ is numerically preserved while count MAE rises from 11.44 to 11.92 and 13.91 and count $R^2$ falls from 0.703 to 0.682 and 0.585, showing that preserved detection AP does not guarantee preserved counting utility. For segmentation, all class IoUs and boundary F1 are reduced by merging, showing that the degradation seems to be uniform across classes. These phenotype-relevant metrics identify where degradation appears, but they do not prove that particular merges crossed a head instance or organ boundary.

\begin{figure}
    \centering
    \includegraphics[width=\linewidth]{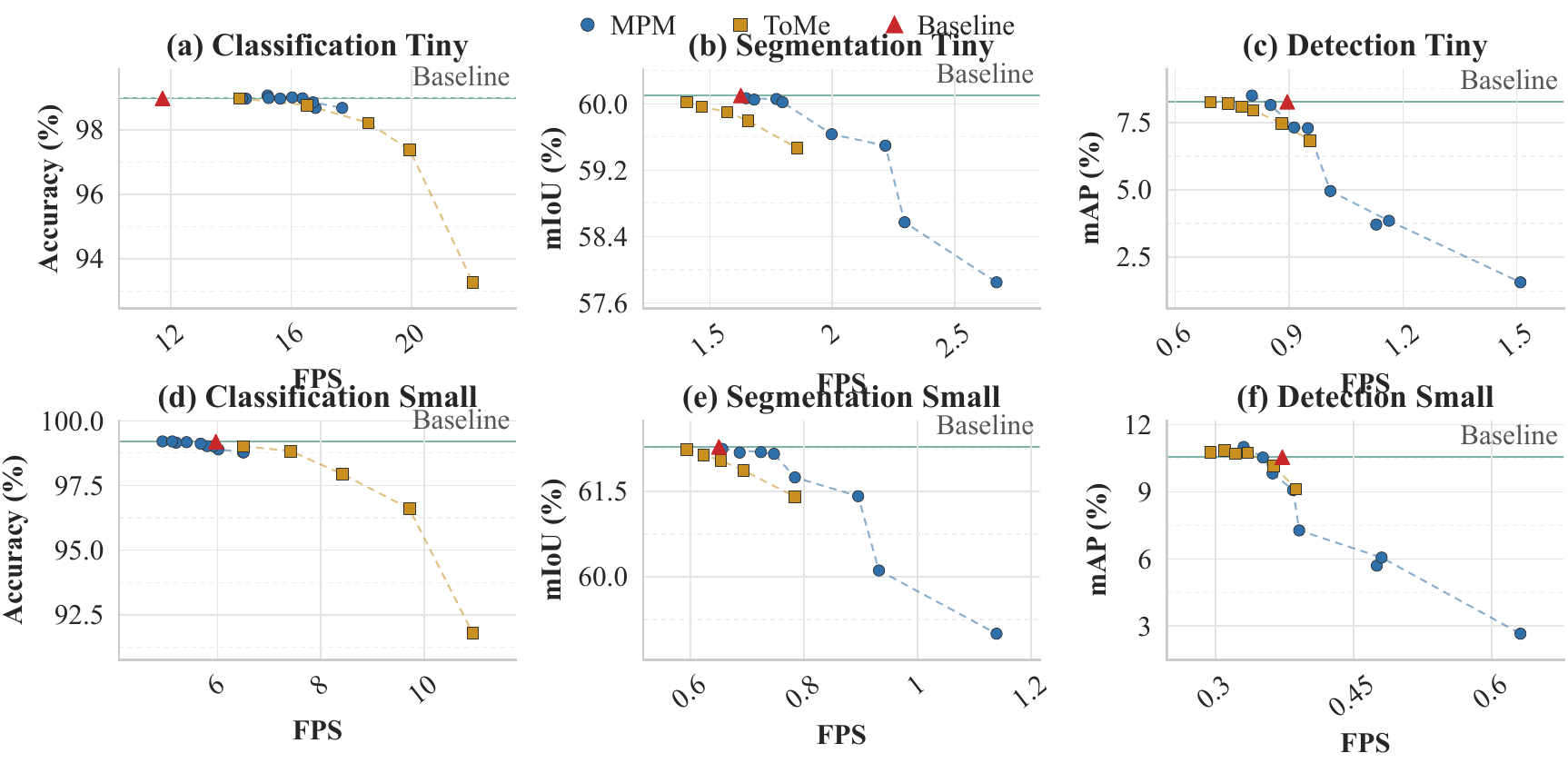}
    \caption{Raspberry Pi 5 batch-one performance--throughput trade-offs on fixed 100-image subsets. The x-axis is FPS and the y-axis is Top-1, mAP, or mIoU. Dashed lines show method-specific Pareto frontiers.}
    \label{fig:locked_three_task_fps_metric_summary_pi5}
\end{figure}

We also report batch-one inference on a Raspberry Pi 5 in \cref{fig:locked_three_task_fps_metric_summary_pi5}. The results are qualitatively similar to those on the GPU: classification is most merge-tolerant, while segmentation and detection degrade faster. Aggressive settings reach about \(2\times\) CPU speedup, but compact detectors have low absolute mAP. We must also note that the device reported thermal throttling.

\begin{table*}[t]
\centering
\small
\setlength{\tabcolsep}{4.0pt}
\caption{Selected Base-backbone points from \cref{fig:locked_three_task_fps_metric_summary}. FPS is measured in images/s. Speedup is the ratio to the within-task Full baseline. Metric denotes Top-1, mIoU, or mAP. $\Delta$ denotes its percentage-point change. Tokens denotes the mean final count per image.}
\label{tab:selected-base-pareto-results}
\resizebox{\linewidth}{!}{
\begin{tabular}{lllrrrrr}
\toprule
Task & Model & Method & FPS & Speedup & Metric (\%) & $\Delta$ pp & Tokens \\
\midrule
Classification & ViT-B/16 & Full & 2743.46 & 1.00 & 99.15 & 0.00 & 197.0 \\
Classification & ViT-B/16 & ToMe \(r=16\), no prop. & 4362.48 & 1.59 & 99.01 & -0.13 & 11.0 \\
Classification & ViT-B/16 & MPM \(L=\{0,2,4,6,8\}\) & 4022.86 & 1.47 & 98.89 & -0.26 & 79.0 \\
\midrule
Segmentation & Segmenter-B/16 & Full & 177.80 & 1.00 & 63.63 & 0.00 & 1025.0 \\
Segmentation & Segmenter-B/16 & ToMe \(r=48\), no prop. & 236.16 & 1.33 & 62.80 & -0.82 & 449.0 \\
Segmentation & Segmenter-B/16 & MPM \(L=\{6,9\}\) & 205.96 & 1.16 & 63.26 & -0.36 & 591.9 \\
\midrule
Detection & YOLOS-B & Full & 5.00 & 1.00 & 25.50 & 0.00 & 1701.0 \\
Detection & YOLOS-B & ToMe \(r=48\) & 5.54 & 1.11 & 26.77 & +1.27 & 1125.0 \\
Detection & YOLOS-B & MPM \(L=\{10\}\) & 5.09 & 1.02 & 26.71 & +1.21 & 1268.1 \\
\bottomrule
\end{tabular}
}
\end{table*}

\begin{table*}[t]
\centering
\small
\setlength{\tabcolsep}{4.2pt}
\caption{Secondary metrics and peak allocated CUDA memory for \cref{tab:selected-base-pareto-results}. IoUs follow Background, Head/Spike, Stem, and Leaf. Boundary F1 excludes Background and uses a two-pixel tolerance.}
\label{tab:selected-secondary-memory}
\resizebox{\linewidth}{!}{%
\begin{tabular}{lllrr}
\toprule
Task & Method & Secondary metric & Value & Peak memory (GB) \\
\midrule
Classification & Full & Macro-F1 (\%) & 99.15 & 0.636 \\
Classification & ToMe $r=16$, no prop. & Macro-F1 (\%) & 99.01 & 0.607 \\
Classification & MPM $L=\{0,2,4,6,8\}$ & Macro-F1 (\%) & 98.89 & 0.614 \\
\addlinespace
Detection & Full & AP$_{50}$ (\%) / count MAE / count $R^2$ & 64.46 / 11.44 / 0.703 & 4.226 \\
Detection & ToMe $r=48$ & AP$_{50}$ (\%) / count MAE / count $R^2$ & 65.20 / 11.92 / 0.682 & 4.655 \\
Detection & MPM $L=\{10\}$ & AP$_{50}$ (\%) / count MAE / count $R^2$ & 65.47 / 13.91 / 0.585 & 4.214 \\
\addlinespace
Segmentation & Full & IoU$_{0/1/2/3}$ / boundary F1 (\%) & 70.14 / 33.33 / 74.71 / 76.32 / 34.88 & 6.023 \\
Segmentation & ToMe $r=48$, no prop. & IoU$_{0/1/2/3}$ / boundary F1 (\%) & 69.06 / 32.52 / 74.00 / 75.63 / 33.79 & 9.142 \\
Segmentation & MPM $L=\{6,9\}$ & IoU$_{0/1/2/3}$ / boundary F1 (\%) & 69.81 / 32.79 / 74.41 / 76.05 / 34.14 & 6.719 \\
\bottomrule
\end{tabular}
}
\end{table*}

\subsection{FlashAttention-2}
\label{sec:flash-attention}

On recent GPUs, FlashAttention-2~\cite{dao2024flashattention2} reduces the attention cost of the unmerged baseline, so token merging no longer guarantees higher throughput. \cref{tab:flash-shared-merging-forward-throughput} reports forward FPS with FlashAttention-2 enabled. For classification, ToMe and aggressive MPM reach only $0.92\times$ and $0.35\times$ baseline throughput. Aggressive MPM improves Segmenter-B/16 throughput by $1.33\times$. Both ToMe ($0.64\times$) and MPM ($0.85\times$) remain slower for tiled YOLOS-B. Once attention is optimized, merge construction, reconstruction, padding, and task-specific pipeline costs can dominate the savings from shorter sequences. Token reduction therefore cannot be treated as backend-independent acceleration.

\begin{table}[t]
\centering
\footnotesize
\setlength{\tabcolsep}{5pt}
\caption{Batch-size-32 bfloat16 forward throughput with FlashAttention-2: median FPS $\pm$ standard error over three runs. Detection aggregates all tiled forward passes per image. ToMe uses no proportional attention.}
\label{tab:flash-shared-merging-forward-throughput}
\begin{tabular}{lllrr}
\toprule
Task & Model & Method & FPS & Speedup \\
\midrule
Cls. & ViT-B/16 & Baseline & \textbf{4047.8 \(\pm\) 160.2} & \textbf{1.00} \\
Cls. & ViT-B/16 & ToMe \(r=24\) & 3723.3 \(\pm\) 147.7 & 0.92 \\
Cls. & ViT-B/16 & MPM \(L=\{0,2,4,6,8\}\) & 1401.6 \(\pm\) 3.5 & 0.35 \\
\addlinespace
Seg. & Segmenter-B/16 & Baseline & 583.2 \(\pm\) 4.3 & 1.00 \\
Seg. & Segmenter-B/16 & ToMe \(r=24\) & 588.2 \(\pm\) 5.1 & 1.01 \\
Seg. & Segmenter-B/16 & MPM \(L=\{0,2,4,6,8\}\) & \textbf{774.6 \(\pm\) 17.3} & \textbf{1.33} \\
\addlinespace
Det. & YOLOS-B & Baseline & \textbf{8.325 \(\pm\) 0.021} & \textbf{1.00} \\
Det. & YOLOS-B & ToMe \(r=24\) & 5.348 \(\pm\) 0.009 & 0.64 \\
Det. & YOLOS-B & MPM \(L=\{0,2,4,6,8\}\) & 7.062 \(\pm\) 0.014 & 0.85 \\
\bottomrule
\end{tabular}
\end{table}

\subsection{Merge Locality}
\label{sec:merge-locality}

\begin{figure}[t]
    \centering
    \includegraphics[width=0.50\linewidth]{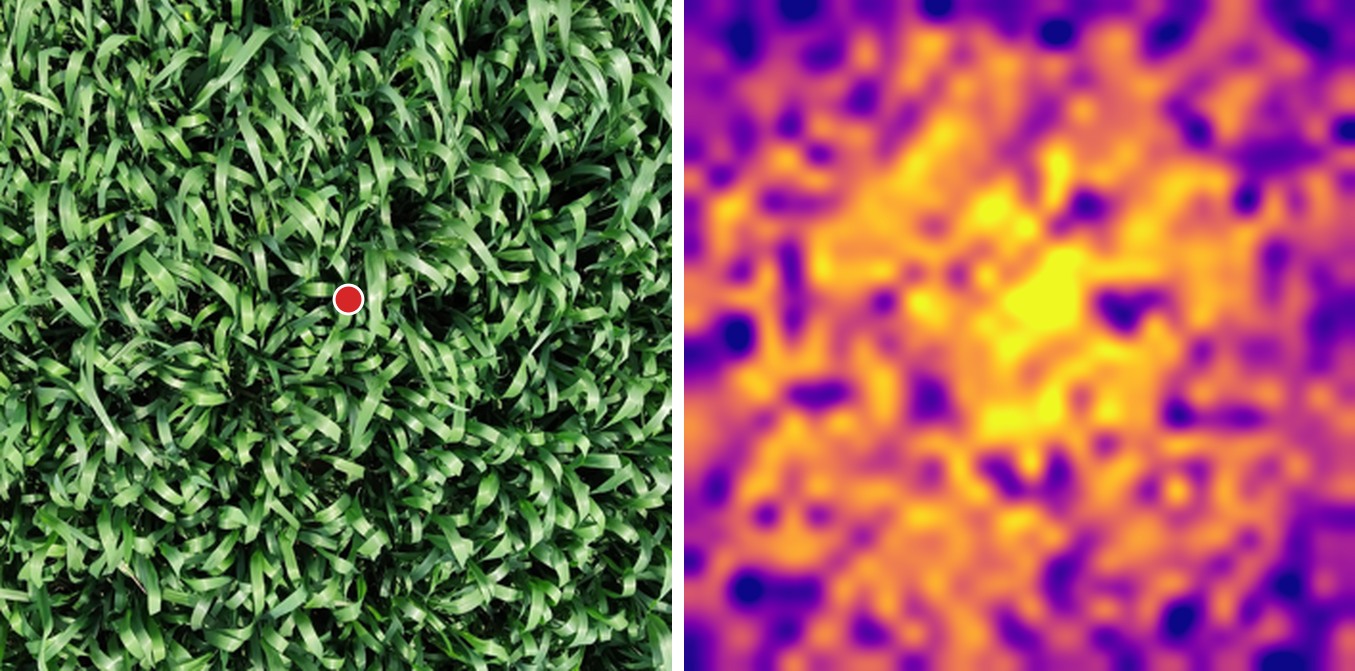}
    \caption{DINOv3~\cite{simeoni2025dinov3} patch-level self-similarity in a wheat canopy image. The red point marks the query patch, warm colors indicate higher cosine similarity at other locations, illustrating repeated visual structure in wheat imagery. This illustration is not a ToMe or MPM merge map and is not used quantitatively.}
    \label{fig:similarity}
\end{figure}

The locality analysis connects token merging to a specific property of plant imagery: self-similarity, visualized in \cref{fig:similarity}. Wheat canopies contain many similar leaves, stems, and heads, so a feature-space merger may find visually plausible pairs that are spatially distant. Neither ToMe nor MPM explicitly enforces spatial locality when selecting token pairs. ToMe forms pairs from attention-key similarity, whereas MPM uses mutual nearest neighbors in feature space. As a result, both methods may merge tokens that are far apart on the image grid if their representations are similar. Whether this is harmful depends on the task. For canopy classification, distant but similar regions can support the same global decision. For wheat-head detection and wheat-organ segmentation, the same operation can blur instance identity and organ boundaries or erase thin organs.

To quantify this behavior, we measure the Manhattan distance between merged token pairs on the patch grid. For representative ToMe and MPM configurations, \cref{tab:merge-distance-locality} reports the mean, median, 95th and 99th percentiles, and the fractions of pairs within distances 2 and 4.

The resulting merge patterns reflect the task structure. In WGSP classification, merged pairs are substantially more spread out: ToMe has median distance 4.29 and only 48.8\% of pairs lie within four steps. The corresponding MPM values are 2.00 and 72.7\%. In GWHD, both methods have median distance 1.00 and at least 97.6\% of pairs lie within four steps. GWFSS is intermediate, although its tail remains broad for ToMe. These distributions are consistent with the canopy-level nature of classification, where similar visual evidence can occur across large regions, while dense tasks preferentially match nearby structures.

\begin{table}[t]
\centering
\small
\setlength{\tabcolsep}{4.5pt}
\caption{Patch-grid Manhattan distances pooled over all merged token pairs for each representative task--method configuration.}
\label{tab:merge-distance-locality}
\begin{tabular}{llrrrrrr}
\toprule
Task & Method & Mean & Median & P95 & P99 & $d\leq2$ (\%) & $d\leq4$ (\%) \\
\midrule
WGSP cls. & ToMe $r=24$ & 5.03 & 4.29 & 12.0 & 16.0 & 28.4 & 48.8 \\
WGSP cls. & MPM $L=\{2\}$ & 3.43 & 2.00 & 11.0 & 14.0 & 56.4 & 72.7 \\
\midrule
GWFSS seg. & ToMe $r=24$ & 3.24 & 1.50 & 13.0 & 29.0 & 69.5 & 84.7 \\
GWFSS seg. & MPM $L=\{2\}$ & 2.23 & 1.00 & 5.0 & 18.0 & 78.6 & 93.7 \\
\midrule
GWHD det. & ToMe $r=24$ & 1.52 & 1.00 & 3.0 & 6.0 & 89.6 & 97.6 \\
GWHD det. & MPM $L=\{2\}$ & 1.31 & 1.00 & 2.0 & 3.0 & 96.3 & 100.0 \\
\bottomrule
\end{tabular}
\end{table}
\subsection{Batch Size and Padding Analysis}
\label{sec:batch-analysis}

MPM's input-dependent compression pads each batch to its longest merged sequence. This cost is absent at batch one but can reduce offline throughput. On 100 fixed batches of 32 test images, \cref{tab:mpm_padding_diagnostic} reports merged-token counts, retained-token distributions, and padding. Aggressive merging averages 20.3, 70.7, and 62.9 padded tokens per classification, segmentation, and detection image, with 95th-percentile costs of 25.5, 91.6, and 86.7. Mean retained count alone is therefore an incomplete efficiency proxy.


\begin{center}
\begin{minipage}{\linewidth}
\centering
\scriptsize
\renewcommand{\arraystretch}{0.90}
\setlength{\abovecaptionskip}{4pt}
\setlength{\belowcaptionskip}{2pt}
\captionsetup{hypcap=false}
\captionof{table}{MPM adaptive-padding on 100 fixed batches of 32 test images, reused across schedules. Pad/img is final padding per image after the last MPM insertion.}
\label{tab:mpm_padding_diagnostic}
\setlength{\tabcolsep}{4pt}
\resizebox{\linewidth}{!}{%
\begin{tabular}{lllrrrrr}
\toprule
Task & Level & MPM layers $L$ & Merged/img &
\multicolumn{2}{c}{Final tokens/img} &
\multicolumn{2}{c}{Pad/img} \\
\cmidrule(lr){5-6}\cmidrule(lr){7-8}
& & & & Median & p95 & Avg. & p95 \\
\midrule
Cls. & Low & $\{10\}$ & 42.0 & 153.0 & 167.0 & 18.9 & 26.4 \\
Cls. & Medium & $\{6,9\}$ & 75.6 & 120.0 & 130.0 & 12.4 & 17.9 \\
Cls. & Aggressive & $\{0,2,4,6,8,10\}$ & 136.8 & 59.0 & 75.0 & 20.3 & 25.5 \\
\addlinespace
Seg. & Low & $\{10\}$ & 238.1 & 782.0 & 824.0 & 58.3 & 81.5 \\
Seg. & Medium & $\{6,9\}$ & 433.4 & 587.0 & 646.0 & 66.7 & 88.3 \\
Seg. & Aggressive & $\{0,2,4,6,8,10\}$ & 768.0 & 250.0 & 311.0 & 70.7 & 91.6 \\
\addlinespace
Det. & Low & $\{10\}$ & 438.1 & 1162.0 & 1182.0 & 26.2 & 37.2 \\
Det. & Medium & $\{6,9\}$ & 734.2 & 866.0 & 895.0 & 38.7 & 54.8 \\
Det. & Aggressive & $\{0,2,4,6,8,10\}$ & 1264.6 & 334.0 & 383.0 & 62.9 & 86.7 \\
\bottomrule
\end{tabular}%
}
\end{minipage}
\end{center}

\FloatBarrier
\section{Discussion and Limitations}

\paragraph{Plant-Specific Interpretation.}
Training-free token merging helps plant phenotyping only when the task, merging schedule, and deployment path align. Growth-stage classification is the clearest case: redundant canopy evidence can be compressed while preserving the global decision. Wheat-head detection and wheat-organ segmentation expose the limits, because repeated heads, thin leaves, stems, and organ boundaries are biological signal rather than mere visual redundancy. The same self-similarity that makes wheat images attractive for compression can therefore make aggressive merging unsafe. The secondary metrics make this limitation concrete: AP$_{50}$ can remain stable while count MAE worsens and count $R^2$ falls, and modest mIoU loss can coexist with lower Head/Spike IoU and boundary F1.

However, these associations do not establish a biological mechanism. The benchmark was not stratified by Zadoks stage, canopy or head density, occlusion, organ scale, perspective, or acquisition height. Locality cannot reveal which semantics a pair connects. Classification tolerance is consistent with global evidence requirements and dense-task fragility with spatial precision, but distance alone does not explain the curves.

\paragraph{Deployment Guidance.}
Token merging should be selected per task and runtime, not enabled by default. On the standard GPU path, classification is the strongest case: ToMe $r=16$ without proportional attention and MPM $L=\{0,2,4,6,8\}$ yield $1.59\times$ and $1.47\times$ speedups with losses below 0.3 Top-1 points. These are measured operating points, not universal loss tolerances.

Segmentation is conditional. ToMe $r=48$ and MPM $L=\{6,9\}$ improve throughput by $1.33\times$ and $1.16\times$, but reduce mIoU, every class IoU, and boundary F1. Region and boundary losses must be checked against the application's tolerance. Detection is the weakest case: mild ToMe preserves mAP only approximately, gives modest speedup, and increases count MAE, while MPM is nearly throughput-neutral. With FlashAttention-2, both tested detection mergers are slower than Full, suggesting that the tiled pipeline is the larger bottleneck.

The same task-level pattern also appears across the tested deployment environments, from the GPU to the Raspberry Pi 5. Measurements show more consistent CPU gains, but the device reported throttling. On every platform, the full baseline should first be profiled with the intended precision, batch size, backend, and complete pipeline. A shorter sequence is useful only if it improves performance along that deployment path.

\paragraph{Scope and Limitations.}
Plain ViT classifiers, Segmenter, and YOLOS isolate sequence behavior. Hierarchical backbones, modern detectors, and foundation-model decoders may differ. We compare two training-free mergers rather than task-trained reduction. Joint training may improve robustness but removes the fixed-checkpoint control.

We also did not profile the energy cost of merging, which may be relevant for edge deployment. Finally, the benchmark is limited to wheat phenotyping. Other crops and plant organs may have different self-similarity and spatial requirements.

\section*{Acknowledgements}
This research was funded by the European Union's Horizon Europe Research and Innovation Programme under the PHENET project, Grant Agreement Number 101094587. This work was granted access to the HPC resources of IDRIS under the allocation 2024-AD010115553 made by GENCI.

\bibliographystyle{splncs04}
\bibliography{main}

\end{document}